\documentclass{article}
\usepackage{spconf,amsmath,graphicx,amsfonts}
\usepackage{array,bbm}
\usepackage{multirow}
\usepackage{xcolor}
\title{LEARNING CASCADED SIAMESE NETWORKS FOR HIGH PERFORMANCE\\VISUAL TRACKING}
\name{Peng Gao, Yipeng Ma, Ruyue Yuan, Liyi Xiao, Fei Wang}%
\address{Department of Electronic and Information Engineering\\Harbin Institute of Technology, Shenzhen}

\begin{document}
%
\maketitle
\begin{abstract}
Visual tracking is one of the most challenging computer vision problems. In order to achieve high performance visual tracking in various negative scenarios, a novel cascaded Siamese network is proposed and developed based on two different deep learning networks: a matching subnetwork and a classification subnetwork. The matching subnetwork is a fully convolutional Siamese network. According to the similarity score between the exemplar image and the candidate image, it aims to search possible object positions and crop scaled candidate patches. The classification subnetwork is designed to further evaluate the cropped candidate patches and determine the optimal tracking results based on the classification score. The matching subnetwork is trained offline and fixed online, while the classification subnetwork performs stochastic gradient descent online to learn more target-specific information. To improve the tracking performance further, an effective classification subnetwork update method based on both similarity and classification scores is utilized for updating the classification subnetwork. Extensive experimental results demonstrate that our proposed approach achieves state-of-the-art performance in recent benchmarks.
\end{abstract}
\section{Introduction}\label{sec1}

Visual tracking is a most fundamental research issue in the field of computer vision, and it is widely developed in numerous applications, such as video surveillance, drone tracking, self-driving vehicle, human-computer interaction, auxiliary medical diagnosis, and many others~\cite{survey2006,survey2014}. Normally, tracking task is to estimate the trajectory of an arbitrary target in an image sequence, given only its initial location at the first frame. Despite the excellent results achieved by numerous tracking approaches~\cite{kcf,mfcmt,mdnet,siamfc,eco,csot} in the past decades, visual tracking is still a challenging problem owing to complicated factors like fast motions, background clutters, motion blurs, deformations, illumination variations, low resolution, occlusions, out of views, scale variations, etc.
\begin{figure}[t]
\centering
\includegraphics[width=\linewidth]{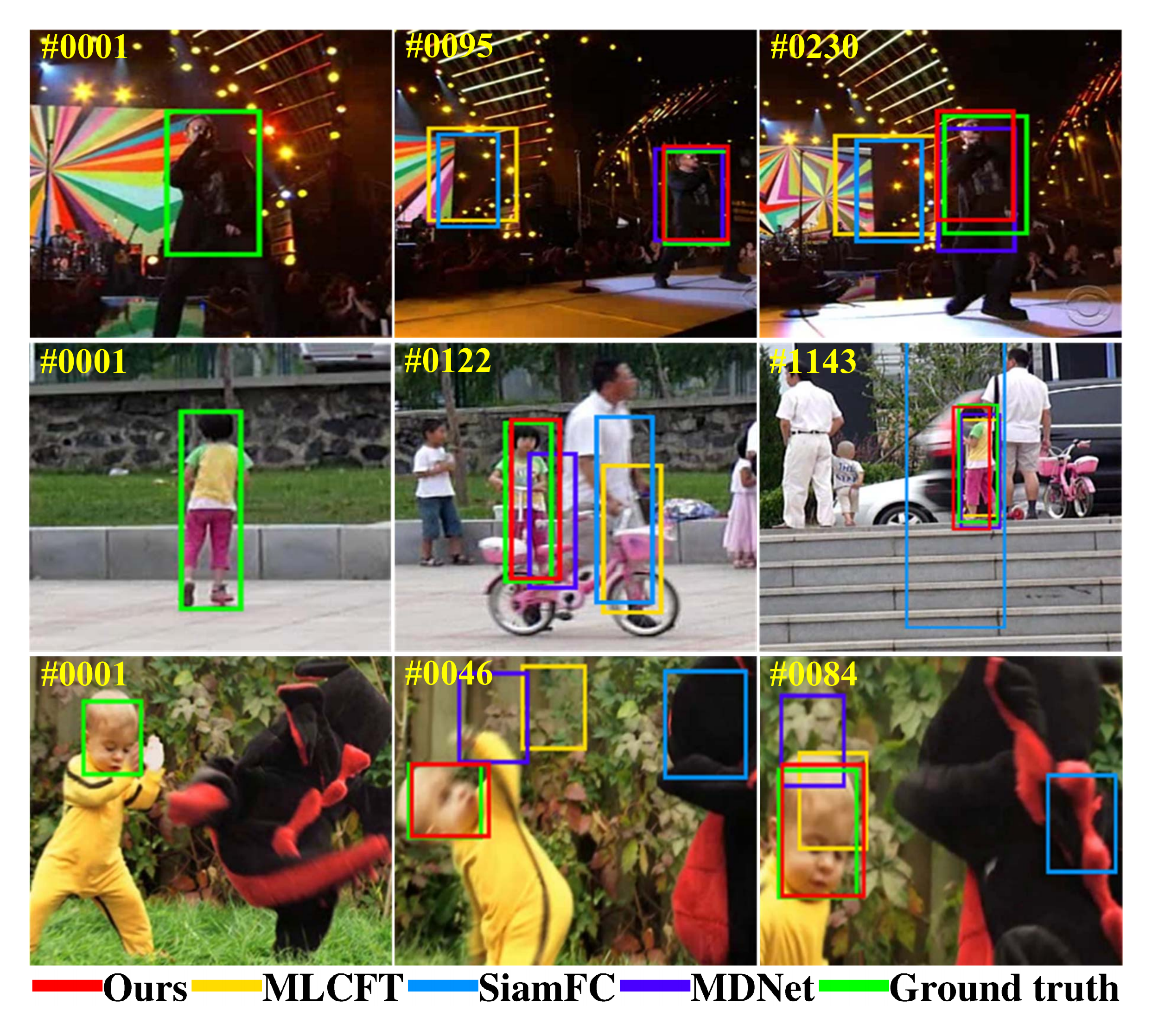}\vspace{-1em}
\caption{Comparison of our proposed tracking approach with three state-of-the-art CNN based trackers: MLCFT~\cite{mlcft}, SiamFC~\cite{siamfc} and MDNet~\cite{mdnet} on three example sequences from OTB2015 benchmark~\cite{otb2015}, respectively. We also present the ground-truth bounding boxes of these example sequences. Best viewed in color.}
\label{fig:1}
\end{figure}
\begin{figure*}[t]
\centering
\includegraphics[width=.95\linewidth]{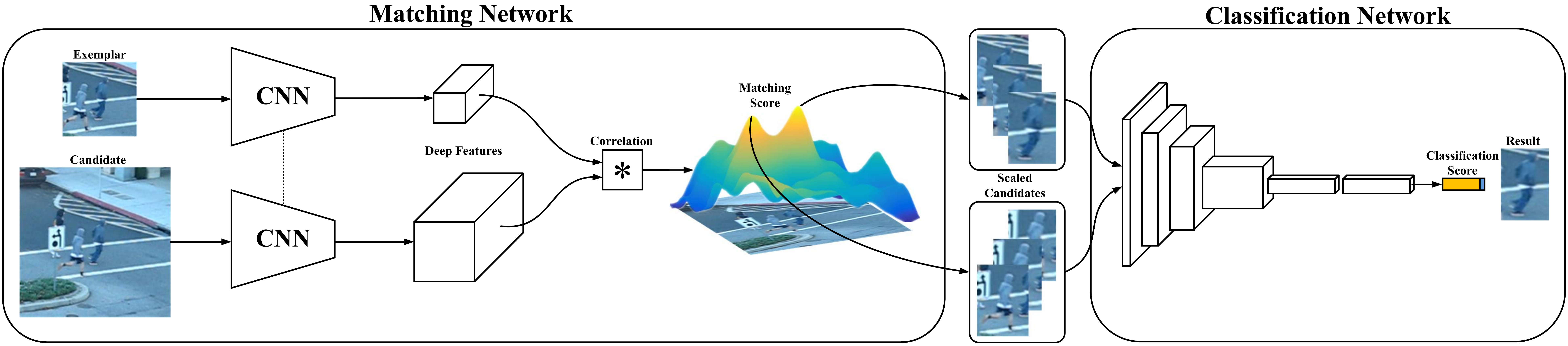}
\caption{The overall framework of our proposed approach.}
\label{fig:2}
\end{figure*}

In recent years, with the tremendous development of deep learning technology~\cite{alexnet,vgg,resnet,senet}, convolutional neural networks (CNN) have attracted increasing attention in the tracking community. Compared with the conventional handcrafted features based trackers~\cite{kcf,mfcmt,ivpai,dsst,staple}, CNN based trackers~\cite{lmsco,sint,ccot,mdnet,satin,cfnet} can easily obtain more competitive tracking performance in multiple benchmarks~\cite{otb2013,otb2015,vot2016}. In general, existing CNN based tracking approaches can be divided into two categories, i.e., matching based trackers and classification based trackers. The former is always pre-trained offline on the video object detection dataset of the ImageNet~\cite{imagenet}. During tracking, it matches the candidates with the exemplar by correlating deep features and does not need online updating. In contrast, the classification based tracking approach transfers a pre-trained network as the classifier and then performs online updating by adding some particular layers~\cite{mdnet}. Although all the CNN based trackers above mentioned have obtained impressive tracking results, there is still great potential to enhance performance further.

In this paper, we propose a novel cascaded Siamese network for high performance visual tracking by integrating both the matching and classification networks. First, a matching subnetwork is exploited to measure the similarity between candidate image and exemplar image and crop scaled candidate patches based on the similarity score. Then, a classification subnetwork which is cascaded with the matching subnetwork learns a target-specific classification scheme online to further determine the optimal tracking results among all scaled candidate patches based on the classification score. Finally, both similarity and classification scores are combined together to indicate whether the classification subnetwork should be updated online or not.

Our main contributions are three folds and summarized as follows:
\begin{itemize}
    \item We propose a novel cascaded Siamese network for high performance visual tracking, which consists of a matching subnetwork and a classification subnetwork.
    \item We utilize an effective model update method to determine the necessity for classification subnetwork online updating.
    \item We conduct extensive experiments on several recent tracking benchmarks, our proposed approach achieves surprisingly good performance both in terms of accuracy and robustness, as shown in Fig.~\ref{fig:1}.
\end{itemize}

\section{Algorithmic Overview}\label{sec31}

The overall framework of our proposed approach is shown in Fig.~\ref{fig:2}. The proposed approach consists of a matching subnetwork for target localization and scaled candidate patches creation and a classification subnetwork for optimal tracking results determination. During the tracking process, an exemplar image \textbf{x} of size $127\times127$ and a candidate image \textbf{z} of size $255\times255$ both centered around the previous position of the target are first fed into the matching subnetwork. The matching subnetwork imitates the fully-convolutional Siamese architecture~\cite{siamfc}, and the similarity between the exemplar image and the candidate image is estimated by calculating the cross-correlation based on their deep features. Then, the possible target positions are chosen by searching the maximum similarity scores, and scaled candidate patches centered at all possible target positions are cropped on the candidate image. Here, the scaling method is similar to that of DSST tracker~\cite{dsst}. Next, the scaled candidate patches are resize to $107\times107$ and classified into foreground or background by the classification subnetwork, and the patch with the highest foreground score will be determined as the optimal tracking result. Finally, we update the classification subnetwork online based on the combination of both similarity and classification score corresponding to the optimal tracking result.

\section{The proposed approach}\label{sec3}

\subsection{Matching Subnetwork}\label{sec32}

In our matching subnetwork, we adopt a fully-convolutional Siamese network which is pre-trained offline with a large video object detection dataset~\cite{imagenet} in an end-to-end manner as the deep feature extractor~\cite{siamfc}. Our aim is to learn a function $f(\textbf{z},\textbf{x})=g(\varphi(\textbf{z}),\varphi(\textbf{x}))$ to compare the exemplar image \textbf{x} with the candidate image \textbf{z} of the same size, where $\varphi(\textbf{z})$ and $\varphi(\textbf{x})$ represent the deep feature maps and $g$ is a similarity metric. We utilize a cross-correlation layer to measure the similarity between the output deep features,
\begin{equation} \label{eq:1}
f(\textbf{z},\textbf{x})=\varphi(\textbf{z})\ast\varphi(\textbf{x})+b \cdot\mathbbm{1}
\end{equation}
where $\ast$ denotes the cross-correlation operation, and $b \cdot\mathbbm{1}$ indicates the bias. Thus, the output $f(\textbf{z},\textbf{x})$ indicates a similarity score map corresponding to the exemplar image compared to the candidate image.

The localization of the target can be estimated at the highest peak on the similarity score map. However, since a video stream always undergoes variations such as fast motion, illumination variation and occlusion, the similarity measurement may be disturbed by similar objects or background noises in the candidate image as shown in Fig.~\ref{fig:2}, and there possibly exist multiple peaks on the similarity score map and the target may locate at one of them. If we estimate the target at wrong peaks, it will leads to inaccurate localization and tracking drift. To solve this problem, we use the classification subnetwork to further determine both the optimal target position and size among all the peaks.

\subsection{Classification Subnetwork}
\begin{figure}[t]
\centering
\includegraphics[width=\linewidth]{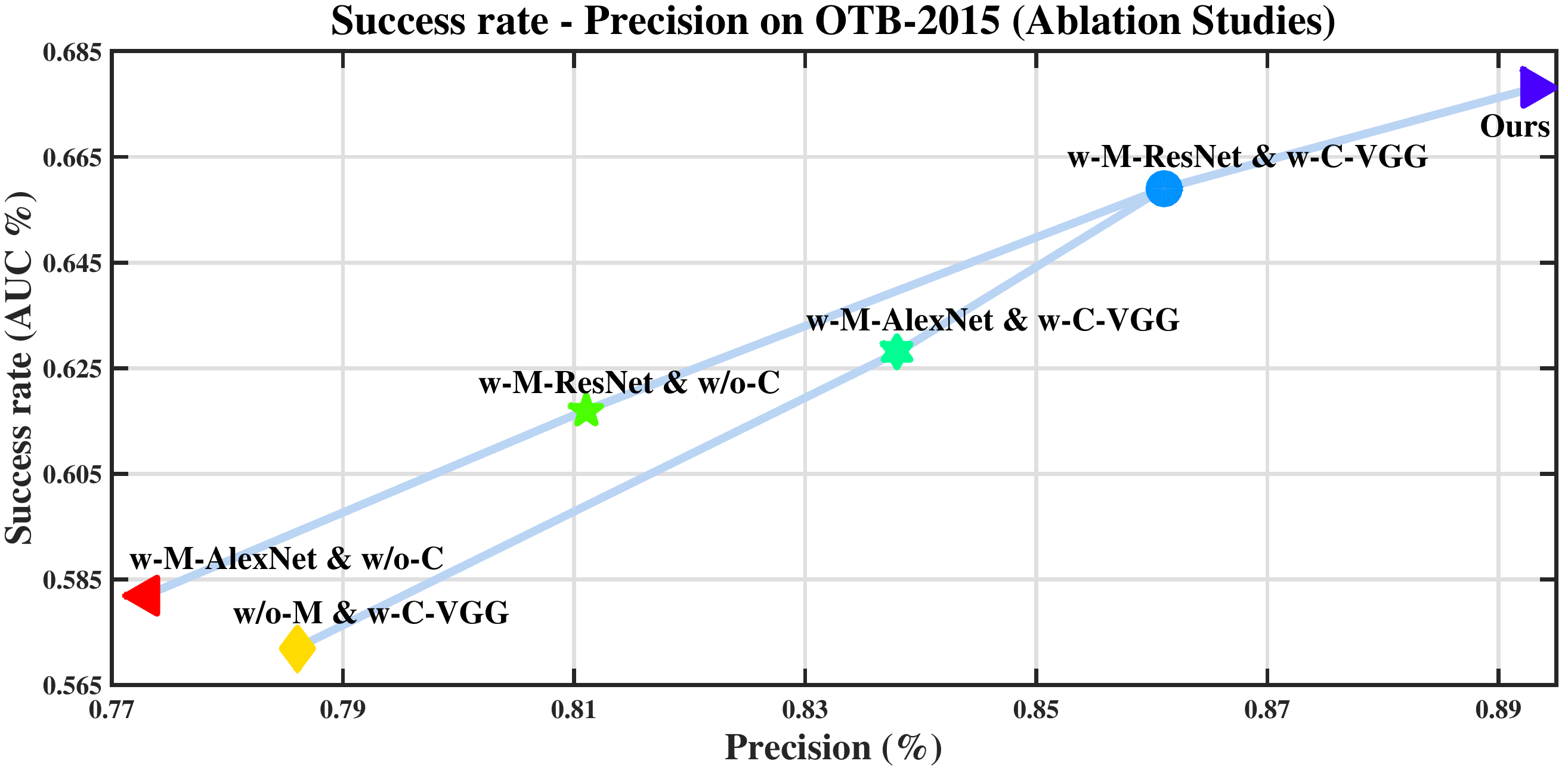}\vspace{-0.7em}
\caption{Ablation studies plot on OTB-2015~\cite{otb2015}. In the legend, \textbf{M} denotes the choice of matching subnetwork and \textbf{C} denotes the choice of classification subnetwork.}
\label{fig:3}
\end{figure}

In Section~\ref{sec32}, we obtain a similarity score map by cross-correlating the output deep features of the feature extractor. Since the similarity score map may not be reliable enough, we treat peaks whose ratio between its score and that of the highest peak exceeding a certain threshold $\gamma_p$ as possible target positions, and the corresponding patches centered at these positions are cropped and scaled as mentioned in Section~\ref{sec31}. After that, a series of scaled candidate patches can be obtained. Thus, we exploit a classification subnetwork for optimal tracking results determination.

The classification subnetwork architecture is similar to that of MDNet~\cite{mdnet} which has three convolutional layers, two fully connected layers and a binary classification layer with softmax cross-entropy loss to output the probabilities of target and background classes, as shown in Fig.~\ref{fig:2}.

Finally, the candidate patch with the highest classification score in the target class will be selected as the optimal tracking result.

\subsection{Updating Method}

During tracking, the parameter of the matching subnetwork are fixed, and all the classification layer and the fully connected layers of the classification subnetwork are fine-tuning online to adapt to variations based on optimal tracking results in the current frame. However, the optimal tracking results are not always reliable for classification subnetwork updates. Inappropriate updates may break down the classification subnetwork due to the ambiguous tracking results.

In order to alleviate this issue, we utilize a simple but effective method for classification subnetwork updating. Assume the similarity and classification scores of current optimal tracking results are $S_M^t$ and $S_C^t$ respectively, and the historical scores of previous $n$ frames are $S_M=\frac{1}{n}\sum_{T=1}^{n}S_M^{t-T}$ and $S_C=\frac{1}{n}\sum_{T=1}^{n}S_C^{t-T}$. If there are no other peaks on the similarity score map that exceed a ratio $\gamma_m$ of the highest peak value, the classification subnetwork will be updated directly based on the current optimal tracking result. In contrast, if there has one or more peaks exceeds the ratio $\gamma_p$ of the highest peak value, we compare both similarity and classification scores with the historical scores. Only when these two scores $S_M^t$ and $S_C^t$ are great than $\gamma_m$ and $\gamma_c$ of their corresponding historical score $S_M$ and $S_C$ respectively, we update the last three layers of our classification subnetwork.

\section{Experiments}\label{sec4}

In this section, we conduct extensive experiments to validate the effectiveness of our proposed cascaded Siamese network. We first detail the implementation of our approach. Then, we investigate the impact of the architecture of the matching and classification subnetworks as well the update method. Finally, we compare our approach with nine state-of-the-art trackers including ECO~\cite{eco}, CCOT~\cite{ccot}, MLCFT~\cite{mlcft}, CACT~\cite{mfcmt}, Staple~\cite{staple}, MDNet~\cite{mdnet}, SiamFC~\cite{siamfc}, KCF~\cite{kcf} and DSST~\cite{dsst} on three tracking benchmarks: OTB-2013~\cite{otb2013}, OTB-2015~\cite{otb2015} and VOT-2016~\cite{vot2016}. The experiments on OTB benchmarks are exploiting two metrics: distance precision and overlap success rate, while the expected average overlap (EAO) is exploited in the VOT dataset.

\subsection{Implementation Details}

\noindent\textbf{Network Architecture.} In the matching subnetwork, we exploit ResNet~\cite{resnet} for deep feature extraction, which followed by a cross-correlation layer. The convolutional layers of the classification network are identical to the corresponding parts of VGG-M~\cite{vgg}, the fully connected layers have 512 output units and the classification layer output 2 scores as described in MDNet~\cite{mdnet}.

\noindent\textbf{Offline Training.} For the training process of both matching and classification subnetworks, sample pairs are selected from the ImageNet video object detection dataset~\cite{imagenet} with random interval. The exemplar and candidate images are picked from the same video. We first load the pre-trained networks to initialize our approach. Then, we apply stochastic gradient descent (SGD) with the learning rate set from $10^{-3}$ to $10^{-4}$ and the momentum of 0.9 to train the networks end-to-end, respectively. More details about the training methods can be found in~\cite{siamfc} and~\cite{mdnet}.

\noindent\textbf{Online Tracking.} During the tracking process, we only update the parameters of the last three layers of the classification subnetwork, and others are fixed. The candidate image is cropped approximately four times the target size centered at the previous position. The certain thresholds $\gamma_p$, $\gamma_m$ and $\gamma_c$ are set to 0.75, 0.8 and 0.6, respectively. The number of historical frames $n$ is set to 6. Moreover, we exploit three scales $1.02^{\{-1,0,1\}}$ to crop candidate pathes at each possible target position.

Our approach is implemented using MXNet~\cite{mxnet} on an Amazon EC2 instance with an Intel Xeon E5 CPU, 61GB RAM and a NVIDIA K80 GPU, 12GB VRAM. It is worth to mention that we retrained MDNet~\cite{mdnet} on ImageNet~\cite{imagenet} since the original MDNet is training with tracking videos that may cause unfair performance over other tracking approaches.
\begin{figure}[t]
\begin{minipage}[b]{1.0\linewidth}
  \centering
  \centerline{\includegraphics[width=\linewidth]{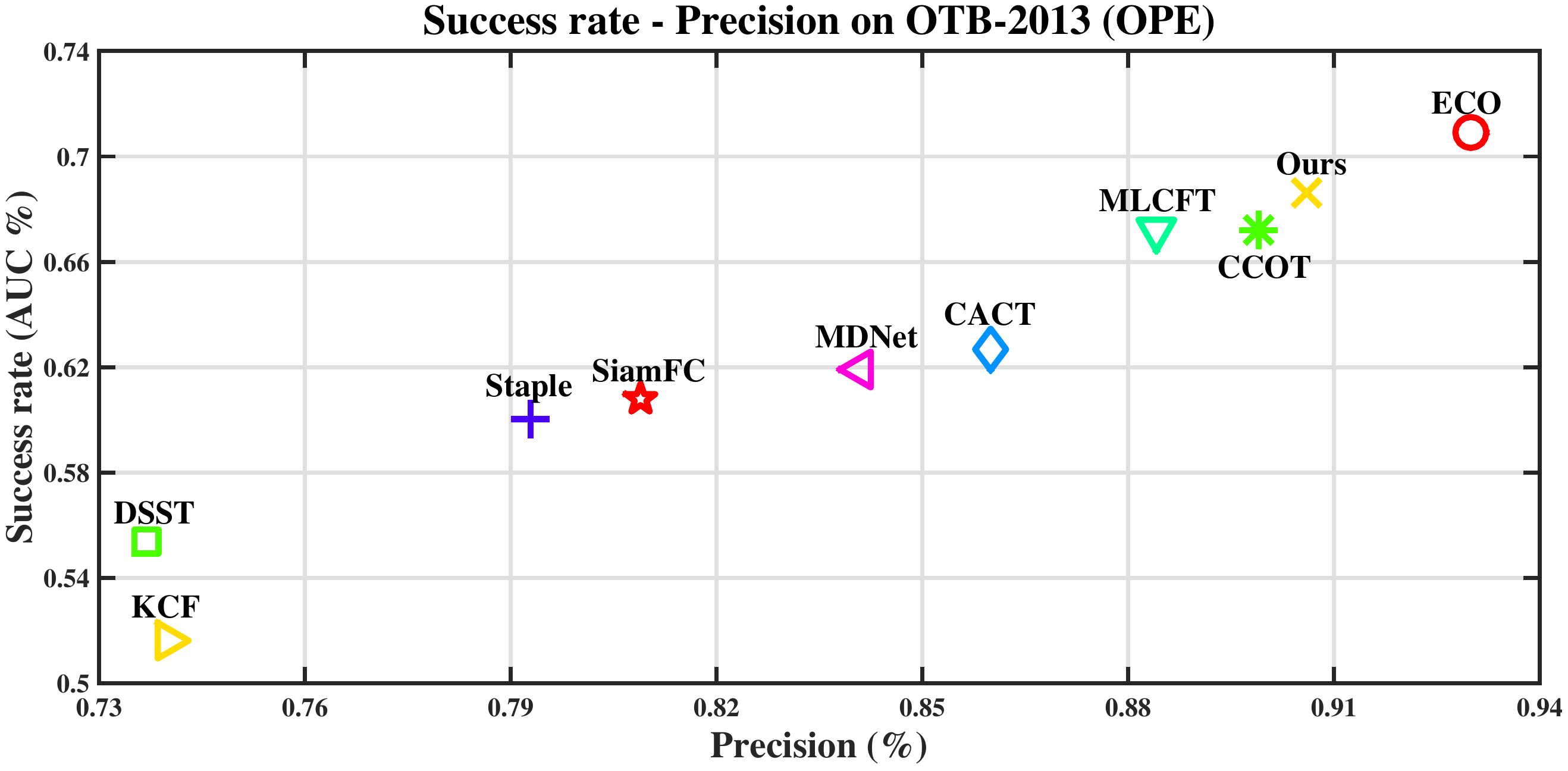}}
  \medskip
\end{minipage}
\begin{minipage}[b]{1.0\linewidth}
  \centering
  \centerline{\includegraphics[width=\linewidth]{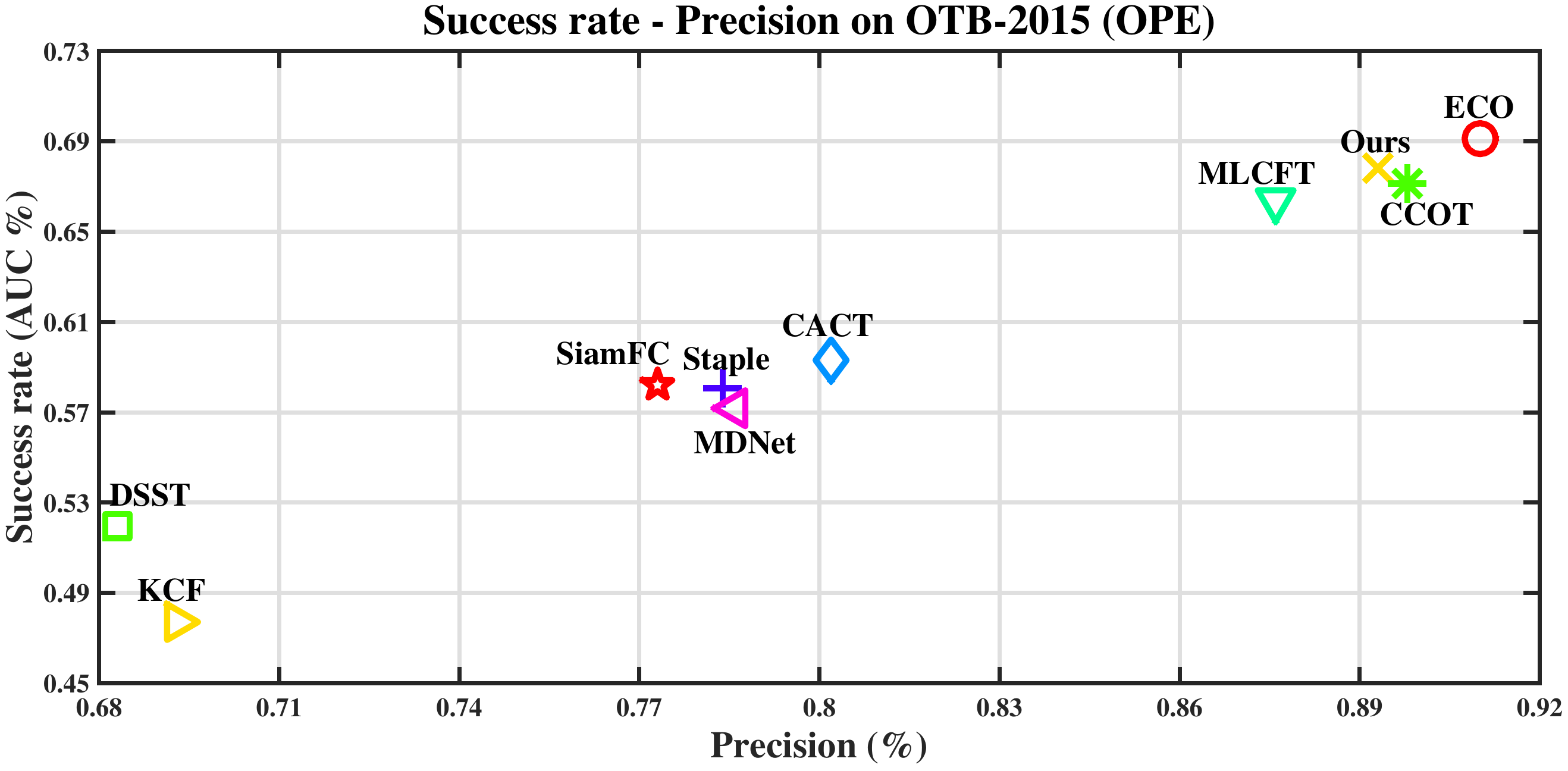}}
  \medskip
\end{minipage}
\vspace{-2.7em}
\caption{Success rate-Precision ranking plots of our approach and nine state-of-the-art trackers on OTB-2013~\cite{otb2013}~(top) and OTB-2015~\cite{otb2015}~(bottom). The better performance a tracker achieves, the closer to the top-right corner of the graph.}
\label{fig:4}
\end{figure}
\begin{figure}[t]
\centering
\begin{minipage}[b]{\linewidth}
  \centering
  \centerline{\includegraphics[width=\linewidth]{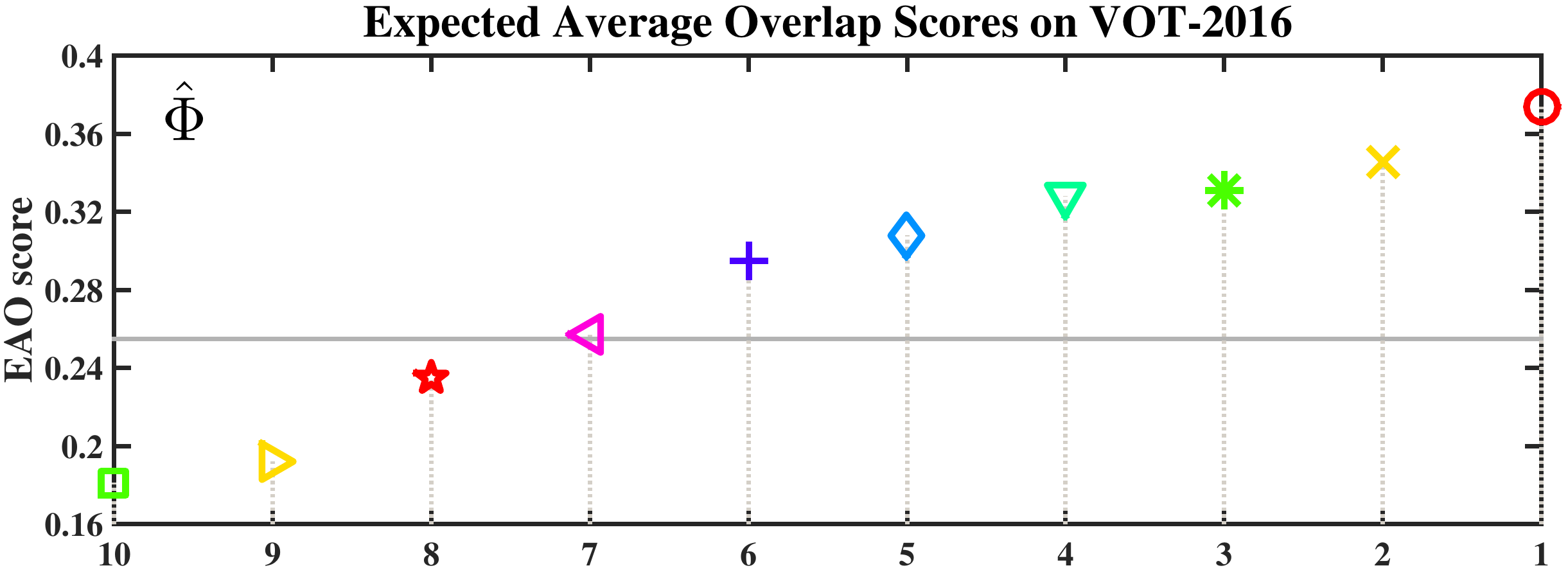}}
  \medskip
\end{minipage}\vspace{-0.2em}
\begin{minipage}[b]{.88\linewidth}
  \centering
  \centerline{\includegraphics[width=.8\linewidth]{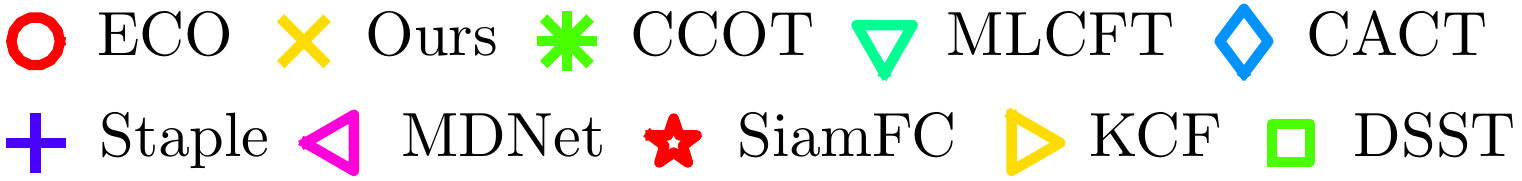}}
  \medskip
\end{minipage}\vspace{-1.5em}
\caption{EAO graph of our approach and nine state-of-the-art trackers on VOT-2016~\cite{vot2016}. The better performance a tracker achieves, the closer to the right of the graph.}
\label{fig:6}
\end{figure}
\subsection{Ablation Studies}

To verify the effectiveness of our designed matching and classification subnetwork as well the update method in our cascaded Siamese network, we conduct ablation studies on OTB-2015 benchmark. The result is shown in Fig.~\ref{fig:3}.

It is clear that the performances of all the variations which are implemented using the components indicated in the plot legend are not as good as our full approach, and each component in our tracking framework is helpful to improve performance. A noteworthy is only our final implementation, denoted by \emph{Ours}, employs the update method.

\subsection{Results on OTB}

We show the success rate-precision ranking plots on OTB-2013 and OTB-2015 benchmarks~\cite{otb2013,otb2015} in Fig.~\ref{fig:4}. It illustrates that the proposed tracker performs better than other re-detection trackers MLCFT and CACT, but is less effective than ECO which exploits continuous convolutional filters.

Overall, our approach attains surprisingly excellent performance both in terms of accuracy and robustness.

\subsection{Results on VOT}

We also evaluate our proposed approach on the VOT-2016 dataset~\cite{vot2016} as shown in Fig.~\ref{fig:6}. The horizontal grey line indicates the state-of-the-art bound according to the VOT committee. Our tracker ranks second in overall performance evaluations based on the EAO measure. Specifically, the performance of our approach excels the CCOT~\cite{ccot} tracker which achieves the best results in the original VOT-2016 challenge.

SiamFC~\cite{siamfc} and MDNet~\cite{mdnet} are the baselines of the proposed approach. Compared to them, our tracker not only learns a matching subnetwork to search the possible target positions, but also benefits from the classification subnetwork to determine the optimal tracking results. What is more, the effective classification subnetwork updating method ensure the robustness of the tracker. Therefore, our cascaded Siamese network outperforms them with a large margin.

\section{Conclusion}\label{sec5}

In this paper, we propose a cascaded Siamese network for high performance visual tracking. Our proposed approach consists of the matching subnetwork for similarity learning and the classification subnetwork for optimal target result determination. Extensive experiments on three recent tracking benchmarks demonstrate competing performance of the proposed tracker over a number of state-of-the-art approaches.

\section{Acknowledgment}\label{sec6}
This work was supported by the National Natural Science Foundation of China under Grant No.~31701187, the Guangdong Provincial Science and Technology Planning Program under Grant No.~2016B090918047, and Promotional Credit from Amazon Web Service, Inc.

\bibliography{icip}

\end{document}